\documentclass[journal]{IEEEtran}
\IEEEoverridecommandlockouts
\usepackage{cite}
\usepackage{amsmath,amssymb,amsfonts,physics}
\usepackage{algorithmic}
\usepackage[ruled,linesnumbered,noend]{algorithm2e}      % microtypography
\usepackage{graphicx}
\usepackage{textcomp}
\usepackage{xcolor}
\def\BibTeX{{\rm B\kern-.05em{\sc i\kern-.025em b}\kern-.08em
    T\kern-.1667em\lower.7ex\hbox{E}\kern-.125emX}}
\usepackage[normalem]{ulem}

\usepackage{subfigure}

\makeatletter
\newcommand{\removelatexerror}{\let\@latex@error\@gobble}
\makeatother

\let\oldnl\nl% Store \nl in \oldnl
\newcommand{\nonl}{\renewcommand{\nl}{\let\nl\oldnl}}% Remove line number for one line
\makeatother

\newcommand{\lA}{\ensuremath{\leftarrow}}

\newcommand{\scr}[1]{\ensuremath{\mathcal{#1}}}\newcommand{\comment}[1]{}

\newtheorem{remark}{\bf Remark}
\SetKwInOut{IP}{I/P}
\SetKwInOut{OP}{O/P}
\SetKwFunction{MORecursivefn}{MO-Recursive-fn}
\begin{document}
%\title{Seasonal Encoder-Decoder Architecture for Multi-step Forecasting}
\title{Seasonal Encoder-Decoder Architecture for  Forecasting}
%\author{\IEEEauthorblockN{Anonymous authors}}
%\author{A1*$^{\S}$,
%       A2$^{\S}$. 
\author{Avinash~Achar*$^{\S}$,
       Soumen~Pachal$^{\S}$. 
        % <-this % stops a space
	\thanks{* Avinash Achar is the corresponding author.}
	\thanks{$^{\S}$ Both authors contributed equally.}
\thanks{ Avinash~Achar, \& Soumen~Pachal   are with TCS Research, Chennai, INDIA.  E-mail: \textit{\{achar.avinash,s.pachal\}@tcs.com.} }
\thanks{Manuscript received }}

\markboth{IEEE TRANSACTIONS }
{Shell \MakeLowercase{\textit{et al.}}: Bare Demo of IEEEtran.cls for IEEE Journals}
\maketitle

\comment{
\author{\IEEEauthorblockN{Soumen Pachal, }
\IEEEauthorblockA{\textit{Research and Innovation} \\
\textit{Tata Consultancy Services}\\
Chennai, India \\
s.pachal@tcs.com}
\and
\IEEEauthorblockN{Avinash Achar}
\IEEEauthorblockA{\textit{Research and Innovation} \\
\textit{Tata Consultancy Services}\\
Chennai, India\\
achar.avinash@tcs.com}
}
}

\begin{abstract}
Deep learning (DL) in general and Recurrent neural networks (RNNs) in particular have seen high success levels in sequence based applications. This paper pertains to RNNs for time series 
	modelling and 
        forecasting.  We
        propose a novel RNN architecture capturing (stochastic) seasonal correlations intelligently while capable of accurate multi-step forecasting. It is motivated from the well-known  encoder-decoder (ED) architecture and multiplicative seasonal auto-regressive model. It incorporates
	multi-step (multi-target) learning even in the presence (or absence) of exogenous inputs. 
        %In essence, the proposed architecture encodes a novel seasonal nonlinear autoregressive model with (or without) exogenous inputs capable of multi-step prediction.  
        It can be employed on single or multiple sequence data. For the multiple
        sequence case, we also propose a novel
        greedy recursive procedure to build (one or more) predictive models across sequences when per-sequence data is less.  We demonstrate via extensive experiments the 
        utility of our proposed
        architecture both in  single sequence and multiple sequence scenarios.

\end{abstract}
%We also propose a particle filter (PF) based approach for approximate optimal prediction. 

\comment{
}
\section{\bf Introduction}
%Travel time is the most sought after traffic information for the travellers, commuters, city planners, traffic agencies etc., especially in congested urban areas, as it is the information which user can perceive. Also, travel time is the most desirable information to develop Intelligent Transportation system applications such as Advanced Traveller Information System(ATIS) and  Advanced Public Transportation Information(APTS). 

Deep recurrent neural networks have seen tremendous success in the last decade across domains like NLP, speech and audio processing \cite{Aaron16}, computer vision \cite{Wang16}, time series 
classification, forecasting and so on.  In particular, it has achieved state-of-art performance (and beyond) in  tasks like handwriting recognition
\cite{Graves09}, speech recognition~\cite{Graves13,Graves14}, machine translation~\cite{Cho14,Sutskever14} and   image captioning~\cite{Kiros14,Xu15}, to name
a few. A salient feature in all these applications that RNNs exploits during learning is the sequential aspect of the data.

In this paper, our focus is on classical time-series, specifically the forecasting issue. The earliest attempts on employing RNNs  for time series forecasting happened 
more than two decades ago \cite{Connor91}, where RNNs were viewed as candidate nonlinear ARMA models. The main advantage of RNNs vs a feed-forward
structure in building non-linear AR models is the weight sharing across time-steps, which keeps the number of weight parameters independent of the order of
auto-regression. The early plain RNN unit was further enhanced with LSTM units \cite{Hochreiter97,Gers00} which increased their ability for long-range
dependencies and mitigated the vanishing gradient problem to an extent. With the exploding, ever-increasing  data (sequential) available across domains
like energy, transportation, retail, finance etc.,  accurate time-series forecasting continues to be an active research area. In particular, RNNs continue to be one of the  widely  preferred predictive modelling tools for time-series forecasting \cite{Fillippo17}. 

\comment{
Prediction in the presence of missing data is an old research problem. Researchers have provided a variety of techniques for data imputation over the years, which can be followed by the predictive modelling step.  In particular, RNNs have also been employed for prediction under missing sequential data over the years. This paper addresses the missing data issue using RNNs in a novel way.
}

Encoder-Decoder (ED) OR Seq2Seq architectures used to map variable length sequences to another variable length sequence were first successfully applied for machine translation \cite{Cho14,Sutskever14} tasks.  From then on, the ED framework has been successfully applied in many other tasks like speech recognition\cite{Liang15}, 
image captioning etc. Given the variable length Seq2Seq mapping ability, ED framework can be naturally utilized for multi-step (target) learning and 
prediction where target vector length can be independent of input vector length. 
%In this paper, we propose two novel interesting variants of the ED framework for time series prediction.

\subsection{\bf Contributions} 
Seasonal ARMAX models \cite{Box90} are generalizations of ARMAX models which additionally capture  stochastic seasonal correlations in the
presence of exogenous inputs. While there are many non-linear variants of  linear ARX or ARMAX models, there is no explicit non-linear
variant of the SARMAX linear statistical model, which is capable of accurate multi-step (target) learning \& prediction to the best of our knowledge.
Existing related DL works either (1) consider  some form of the ED approach for multi-step time-series prediction with exogenous inputs, 
without incorporating stochastic seasonal correlations\cite{Wen17}, (2) many-to-many RNN architecture with poor multi-step predictive ability \cite{Flunkert17,DeepStateSpace18}, (3) incorporate stochastic seasonal correlations using additional skip connections  via 
a non-ED  predictive approach  \cite{Lai18} to emphasize correlations from one cycle (period) behind, (4) capture  deterministic seasonality (additive 
periodic components) in  time-series using a Fourier basis in a overall feed-forward DL framework \cite{Oreshkin20}.

Our main contribution here
involves a novel ED architecture for seasonal modelling using more than one encoder and incorporating multi-step (target) learning.
 %The ED framework enables a direct multi-target (step) learning which is not feasible in most earlier RNN architectures. 
 The overall  contribution summary is as follows:
\begin{itemize}
	%\item  We compute the extent (or order) of spatial dependence a given section travel time might experience from its previous sections using ideas from time-series analyis.	
	\item  We propose a novel Encoder-Decoder based nonlinear SARX model which explicitly incorporates (stochastic) seasonal correlations. The framework can elegantly address 
		multi-step (target) learning with exogenous inputs. It allows for multiple encoders depending on the order of seasonality. {\em The seasonal
		lag inputs are intelligently split between encoder and decoder inputs based on idea of the multiplicative seasonal AR model.} 
		 	
			\comment{
	\item We propose a novel ED based learning scheme for time-series prediction in presence of {\em missing data}.  The missingness pattern in the
		input window is encoded as it is (without imputation) using two encoders with variable length inputs. The resulting encoding is lossless
		while the compression can be substantial depending on the structure of the missingness pattern. 	}
	
	\item  To utilize the above novel scheme for multiple sequence data, where per sequence data or variation in exogenous variables is less, we propose a novel greedy recursive procedure by
		grouping normalized sequences. The idea is to build one or a few background models which can be used to predict for all sequences.
	
	\item We demonstrate effectiveness of  the  proposed architecture on real data sets involving both single and multiple sequences. 
	%We   (i) real demand data from the Australian energy market 	(ii) ???real sales data from multiple sources, where the data comes as multiple sequences.  
Our experiments illustrate  the proposed method's 
performance is competitive with state-of-art and outperforms some of the existing methods.  
\end{itemize}

The paper overall is organized as follows. Sec.~\ref{sec:RW} discusses related work and puts the proposed work in perspective of current literature. Sec.~\ref{sec:Methodology}
describes the proposed seasonal architecture for a single time-series. We next explain a novel recursive grouping algorithm in Sec.~\ref{sec:GreedyRec}, which allows the proposed 
seasonal architecture to handle multiple 
sequence data. By bench-marking against various state-of-art baselines, we demonstrate the efficacy of our proposed architecture on both single and multiple sequence scenarios in
Sec.~\ref{sec:Results}. We provide concluding remarks in Sec.~\ref{sec:Conc}.

\section{\bf Related Work}
\label{sec:RW}
Times series forecasting has a long literature spanning more than five decades. Classical approaches  include AR, MA, ARMA\cite{Box90}, exponential smoothing, linear state space models etc. A wide spectrum of non-linear approaches have also been explored over the years ranging from feed-forward networks, RNNs \cite{Yagmur17,Connor91}, SVMs\cite{SVM09}, random forests \cite{RF14} and so on to the recent temporal convolutional networks (TCN) \cite{TCN18},  with applications spanning across domains. These non-linear techniques have been shown to outperform the traditional techniques. The renewed surge in ANN research over the past decade has seen DL in particular being significantly explored in time series forecasting as well. For a review on deep networks for time series modelling, please refer to \cite{Langkvist14}.

While traditionally time-series forecasting has focused on single time-series or a bunch of independent sequences, the problem of simultaneously forecasting a huge set of correlated  time series is gaining traction given the increased availability of such data.
Examples  include demand forecasting of items in retail, price prediction of stocks, traffic state/congestion across signals, weather patterns across locations etc. There exist many recent sophisticated approaches tackling this high-dimensional problem \cite{Sen19,TRMF16,Flunkert17,DeepStateSpace18,Wen17,Lai18}.
Of these, \cite{Sen19,TRMF16} adopt some variant of a matrix/tensor factorization (MF) approach on the multi-variate data. \cite{Lai18} employs  CNN layers first (on the 2-D data) followed by RNN layers.  
\cite{Flunkert17,DeepStateSpace18,Wen17} consider an RNN architecture at a single time-series level and extend it to multiple time series by essentially scaling  sequences. 
Our proposed architecture is also at a single time-series level but different due to the seasonality feature. 
Our approach to handle multi-sequence learning also employs sequence-specific scaling but goes much beyond this, as described in Sec.~\ref{sec:GreedyRec}.

\subsection{\bf Proposed architecture in perspective}
\label{sec:SeasEDSurvey}
 Two uni-variate ED based attention variants   and simple multivariate extensions of these (which do not look scalable to large number of 
 sequences)   is considered in \cite{Yagmur17}. 
% While attention layer is necessary for NLP applications like machine translation, its need from a time-series perspective is debatable.
 \cite{Yagmur17} doesn't consider exogenous inputs in their architecture. 
To incorporate seasonal correlations especially when  period or cycle length is large, it
would need  a proportionately large input window width  depending on order of seasonality. This would need a large
proportionate set of additional parameters to be learnt which capture the position based attention feature of \cite{Yagmur17}.
This can also lead to (i)processing irrelevant inputs which may not influence prediction (ii)vanishing gradient issue in-spite of attention.  While in our method, we consider multiple encoders where the $k^{th}$ encoder captures the correlations from exactly $k$ periods behind the current instant.
{\em By picking the state from the last time-step of each of these encoders and concatenating them, there is equal emphasis/representation from each of the cycles towards the decoder input.}
This is also unlike \cite{Yagmur17}, where  a convex combination of all
states (of the single encoder) is the context vector. This may not retain information distinctly from each seasonal cycle for instance.
Also, in our approach, since we  split inputs across cycles into
parallel encoders each of  length  much lesser compared to a single encoder, the vanishing gradient issue is potentially better mitigated.

\comment{
 The context vector from the encoder that is fed at every step of the decoder is different in the presence of an attention layer.
 The context vector is a convex combination of the state vectors at each time-step of the decoder.   They introduce and learn  an additional weight or emphasis vector to   s
  }

  \cite{Wen17}  propose an ED architecture where  targets are fed at  decoder output, while  exogenous inputs at  prediction instants are fed as
decoder inputs.
%It does a multi-step (target) learning at the decoder unlike most existing approaches.
%which adopt a recursive application of one-step prediction  for multi-step prediction.
{\em Our seasonal ED architecture can be viewed as a non-trivial generalization of this Seq2Seq architecture  for multi-step prediction.}
They also suggest certain improvements to the basic ED architecture for quicker learning etc. They also consider probabilistic (or interval forecasts) using quantile loss, which can be readily incorporated in our proposed architecture as well.

DeepAR \cite{Flunkert17} and DeepSS \cite{DeepStateSpace18}  use a many-to-many architecture
with exogenous inputs (NARX model) for sales prediction of multiple amazon products). They don't consider
stochastic seasonal correlations.
%The underlying RNN workhorse is a many-to-many architecture, which at each step feeds the current exogenous variable (price for instance)  and the previous sales as input, while the current sales is the target.
During multi-step
prediction, the prediction of the previous step is recursively fed as input to the next time-step, which  means this strategy can lead to recursive error accumulation.
%This architecture can also be viewed as a restricted ED where the decoder and encoder share the same structure and weights.
Both methods also consider probabilistic forecasts by modelling network outputs as parameters of a negative binomial OR a linear dynamical system.

\cite{Lai18} propose LSTNet, another multiple time-series approach where a combination of CNN and RNN approaches are employed. The convolutional filters filter in only one dimension (namely time) across the 2-D input time-window  to learn cross-correlations across sequences. The subsequent RNN part attempts to capture
stochastic seasonal correlations via skip connections from a cycle (or period) behind.  Applications where  period is large results in unusually long input windows. 
%This can kick in the vanishing gradient problem and render large training times. 
While in our approach, we avoid skip connections and feed the seasonal lags from each cycle into a separate encoder (one or more depending on the order of seasonality).
  %The context vectors from these multiple encoders are appended and the merged context vector is further fed as a start state (and possibly as inputs) to thedecoder.

  N-Beats \cite{Oreshkin20} considers a DL approach using feed-forward structures where  deterministic periodicities (referred to seasonality as well in  literature) are captured using Fourier basis. A signal in general can exhibit both kinds of seasonality: (i)deterministic periodic components (ii)stochastic seasonal correlations. A linear seasonal ARMA model for instance only captures stochastic seasonal correlations which is different from  additive
  deterministic periodicity. Our seasonal ED architecture  precisely captures such stochastic correlations  making it distinct from N-Beats.

 Earlier described MF approaches though capture global features in the multi time-series setting, either do
not (1)capture seasonality, (2)not allow for predictions with exogenous inputs or (3)learn with multi-step targets as in our approach. For  details on MF methods, refer to appendix.

\section{\bf Proposed Seasonal ED Architecture}
\label{sec:Methodology}

%We describe all aspects of our technical contribution in this section. In particular, 
Sec.~\ref{sec:Seas} and \ref{sec:EDMult} respectively describe the motivation (from the linear multiplicative seasonal model) and actual proposed RNN architecture
capturing a seasonal NARX
model capable of multi-step prediction. 
Sec.~\ref{sec:GreedyRec} describes a heuristic algorithm to adapt the proposed architecture for multiple sequence prediction.
%could be leveraged when data is in the form of multiple sequences of short length.

Amongst  three standard recurrent structure choices of plain RNN (without gating), LSTM~\cite{Hochreiter97} and GRU~\cite{Chung14}, we choose  GRU
in this paper. Like  LSTM unit,  GRU also has a gating mechanism to mitigate vanishing gradients and have more persistent memory. But  
lesser gate count in GRU keeps   number of weight parameters much smaller. GRU unit as the building block for RNNs is currently 
ubiquitous across sequence prediction
applications \cite{Gupta17,Ravanelli18,Che16,Nicole20}.
A single hidden layer plain RNN unit's hidden state would be
%\vspace{-0.1in}
\begin{equation}
	h_t = \sigma(W^{h} h_{t-1} + W^{u} u_t)
\end{equation}
where $W^{h}$, $W_{u}$ are the weight matrices associated with the state at the previous time-instant $h_t$ and the current input ($u(t)$) respectively,
$\sigma(.)$ denotes sigma function.
GRU based cell computes its hidden state (for one layer as follows)
%\vspace{-0.1in}
\begin{eqnarray}
	z_t & = &\sigma(W^z u_t + U^z h_{t-1})  \\
	r_t & = & \sigma(W^r u_t + U^r h_{t-1}) \\
	\tilde{h}_t & = &tanh(r_t \circ U h_{t-1}  + W u_t) \\
	h_{t} & = & z_t \circ h_t + (1 - z_t)\circ \tilde{h}_t
\end{eqnarray}
where $z_t$ is update gate vector and $r_t$ is the reset gate vector. If the two gates were absent, we essentially have the plain RNN. $\tilde{h}_t$ is
the new memory (summary of all inputs so far) which is a function of $u_t$ and $h_{t-1}$, the previous hidden state. The reset signal controls the
influence of the previous state on the new memory. The final current hidden state is a convex combination (controlled by $z_t$) of the new memory and the memory at the previous
step, $h_{t-1}$. All associated weights $W^z$, $W^r$, $W$, $U^z$, $U^r$, $U$ are trained using back-propagation through time (BPTT).    
\begin{figure*}[!thbp]
\center
%\includegraphics{../../SIAM/SDMFiles/PlotsSDM/AAAI19FigData/EKFVsPFMAPE.pdf}
%\begin{tabular}{cc}
 %\includegraphics[width=7.0in, height=6.0in]{../../../../SampleFiguresInkScape/Retail/SeasonalityMultiStepPaper.png}
 \includegraphics[width=7.0in, height=3.95in]{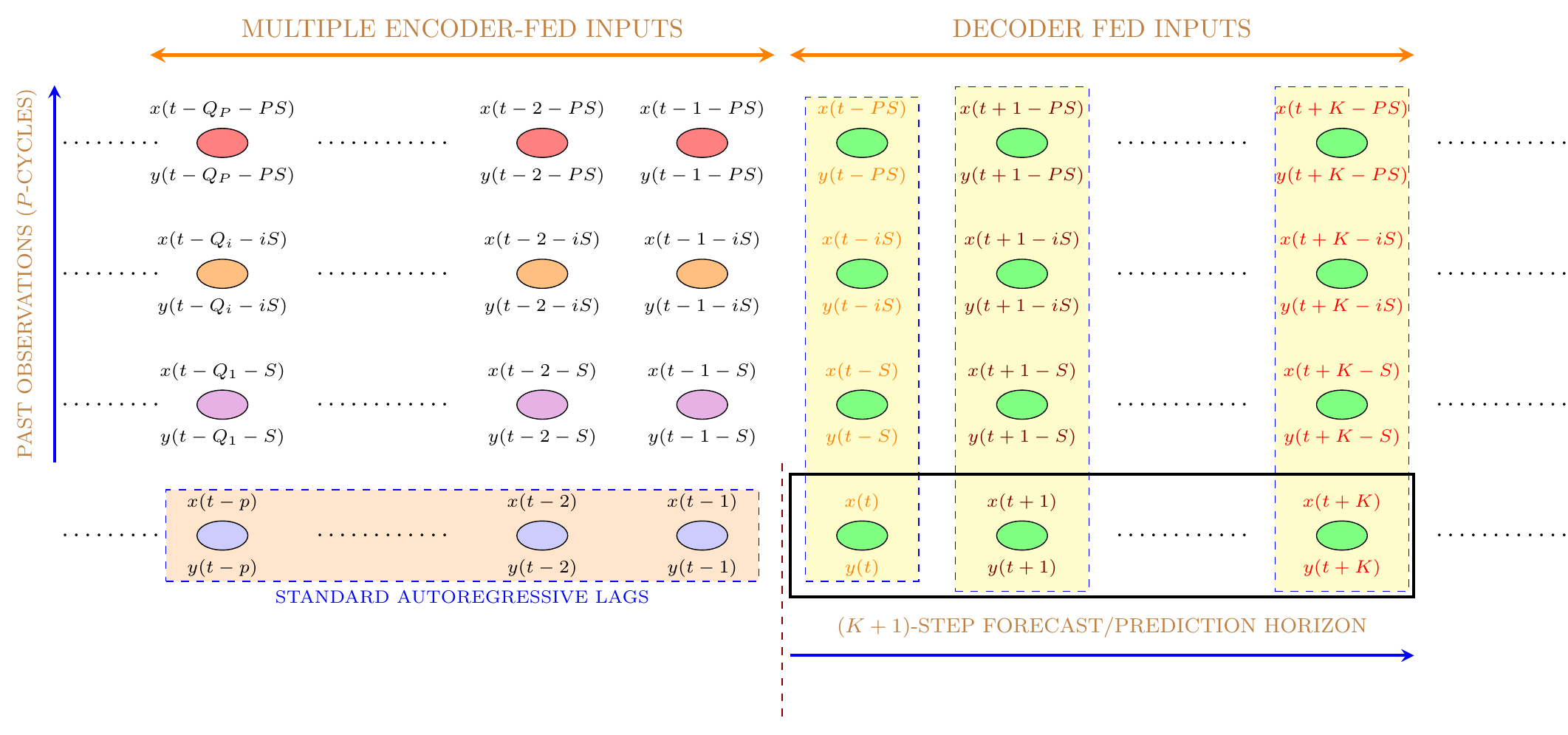}
%\subfigure[RunTime Comparison]{\includegraphics[width=1.6in,height=1.2in]{../SIAM/SDMFiles/PlotsSDM/AAAI19FigData/EKFVsPFRunTime.pdf} \label{fig:RunTimeEKFvsPF}}
%\end{tabular}
\caption{Single sequence which starts from top left and ends on bottom right. Illustration of the Multi-step Seasonal NARX model input-outputs.}
\label{fig:SeasMS}
%\vspace{-0.10in}
\end{figure*}

\subsection{\bf Motivation from multiplicative SAR model}
\label{sec:Seas}
{\bf } The proposed {\em ED-based} multi-step, seasonal-NARX architecture can be  motivated from the classical linear multiplicative SAR model as
described in this section.  A
multiplicative
seasonal AR  model \cite{brock:11} is a stochastic process which satisfies the following equation.
\begin{equation}
%(1-\psi_1L - \psi_2L^2 - \dots -\psi_pL^p)(1-\Psi_1L^{\scr{S}} - \Psi_2L^{2\scr{S}} - \dots -\Psi_kL^{P\scr{S}})y(t) =  e(t).
(1-\psi_1L  - \dots -\psi_pL^p)(1-\Psi_1L^{\scr{S}} -  \dots -\Psi_kL^{P\scr{S}})y(t) =  e(t).
\label{eq:MSAR}
\end{equation}
where, $e(t)$ is a zero mean, white noise process with unknown variance, $\sigma_e^2$. $L^p$ is the one-step delay operator applied $p$ times i.e. $L^p{y(t)} =
y(t-p)$. In a multiplicative SAR model (\ref{eq:MSAR}), the auto-regressive term is a product of two lag polynomials: (a) first capturing  standard lags of 
order up to $p$, (b) second capturing  influence of  seasonal lags at multiples of the period $\scr{S}$ and order up to $P$. Let us expand the associated product in eqn.~(\ref{eq:MSAR}) to obtain  
\begin{align}
	y(t)= & { a_1y(t-1)+a_2y(t-2)+\dots+ a_py(t-p)} + { b_0^{1}y(t-\scr{S})}\nonumber \\
	& {+b_1^{1}y(t-\scr{S}-1) +\dots+b_{p}^{1}y(t-\scr{S}-p)} + \cdots\,\cdots+ \nonumber \\ & {
	b_0^{P}y(t-P\scr{S})+b_1^{P}y(t-P\scr{S}-1) +\dots+}   \nonumber \\ &
	{ b_{p}^{P}y(t-P\scr{S}-p)} + e(t)
    \label{eq:ASAR}
\end{align}
Note in the above equations $p$ is assumed significantly less than $S$.  Expanding out the product of the two polynomials in
eqn.~(\ref{eq:MSAR}) and comparing it with eqn.~(\ref{eq:ASAR}), yields the following relations between the respective coefficients: $a_i = \psi_i$, $b_0^1 = \Psi_1$, $b_i^1 = -\psi_i \Psi_1$, $b_0^{k} =
\Psi_k$,
$b_i^{k} = -\psi_i \Psi_k$. 
{\em Observe from eqn.~\ref{eq:ASAR} that $y(t)$ is linearly auto-regressed w.r.t three types (categories) of inputs: (a) its $p$ previous values ($y(t-1),y(t-2) \dots y(t-p)$) (b) values exactly a 
period $\scr{S}$ (or an integral multiple of $\scr{S}$ lags) behind, up to $P$ cycles $\left(y(t-S), y(t-2S)\dots y(t-PS)\right)$ (c) $P$
groups of $p$ consecutive values, where $i^{th}$ such group is immediately 
previous to $y(t-iS)$, where $i=1,2,\dots P$. For instance, $y(t-iS-1), y(t-iS-2),\dots y(t-iS-p)$ is the $i^{th}$ group of these $P$ groups}. 

{\em We generalize the above structure of the expanded  SAR model as follows.} We first allow all co-efficients in
eqn.~(\ref{eq:ASAR}) to be unconstrained. We further assume that the $P$ groups of consecutive values
(indicated in (c) above) need not be of the {\em same} size $p$. The below equation demonstrates this.
%generalization. 
\begin{align}
    y(t)= & { a_1y(t-1)+\dots+ a_py(t-p)} + { b_0^1y(t-\scr{S})}\nonumber \\
    & {+b_1^1y(t-\scr{S}-1) +\dots+b_{Q_1}^1y(t-\scr{S}-Q_1)} + \cdots\,\cdots+ \nonumber \\ & {
	b_0^{P}y(t-P\scr{S})+b_1^{P}y(t-P\scr{S}-1) +\dots+}   \nonumber \\ &
	{ b_{Q_{P}}^{P}y(t-P\scr{S}-Q_{P})} + e(t)
    \label{eq:ASARGen}
\end{align}
Please note that $Q_{i}$ denotes the size of the $i^{th}$ such group. The RNN structure we propose here adopts a nonlinear auto-regressive version 
of eqn.~(\ref{eq:ASARGen}) given as follows.
\begin{align}
	y(t)= & { F\left(\dashuline{y(t-1),y(t-2),\dots\dots\dots\dots\dots, y(t-p)},\right.} \nonumber \\
	& { \underbrace{y(t-S), y(t-2S)\dots\dots\dots\dots\dots y(t-PS)}, } \nonumber \\ 
	& {\underline{y(t-\scr{S}-1),y(t-\scr{S}-2),\dots\dots\dots, y(t-\scr{S}-Q_1)}, } \nonumber \\
	& {\underline{y(t-2\scr{S}-1),y(t-2\scr{S}-2),\dots, y(t-2\scr{S}-Q_2)},\dots	 }   \nonumber \\ 
	& {\left. \underline{y(t-P\scr{S}-1),y(t-P\scr{S}-2),\dots,y(t-P\scr{S}-Q_{P})}\right) } 
    \label{eq:ASNARGen}
\end{align}
Note that the $3$ categories above indicated by the $3$ different styles of underlining correspond to the three categories ((a),(b),(c)) described 
earlier. 
{\em In the 
presence of an additional exogenous variable (process) $x(t)$, we additionally regress the endogenous variable
$y(t)$ w.r.t to $x(t)$ at the current time $t$  and all those previous time instants where $y(t)$ is exactly auto-regressed as per eqn.~(\ref{eq:ASARGen}) 
as follows.}
\begin{align}
	y(t)=&{F\left(x(t),\dashuline{x(t-1),y(t-1),\dots\dots\dots, x(t-p),y(t-p)},\right.} \nonumber \\
	& { \underbrace{x(t-S),y(t-S),\dots\dots\dots,x(t-PS),y(t-PS)}, } \nonumber \\ 
	& {\underline{x(t-\scr{S}-1),y(t-\scr{S}-1),\dots,y(t-\scr{S}-Q_1)}, } \nonumber \\
	& {\underline{x(t-2\scr{S}-1),y(t-2\scr{S}-1),\dots,y(t-2\scr{S}-Q_2)},\dots }   \nonumber \\ 
	& {\left. \underline{x(t-P\scr{S}-1),y(t-P\scr{S}-1),\dots,y(t-P\scr{S}-Q_{P})}\right) } 
    \label{eq:ASNARXGen}
\vspace{-0.10in}
\end{align}
The above model could be used for multi-step prediction by recursively computing single-step predictions sequentially. However, this
 can accumulate errors and lead to poor performance.
Instead of predicting $y(t)$ alone,  to predict $y(t)$ and $y(t+1)$ simultaneously, by the grammar of the multiplicative SARX model (given
data till $t-1$), we 
potentially need {\em additional inputs} for prediction. Being a model with exogenous inputs, it definitely needs $x(t+1)$ additionally. Being 
a seasonal model, it would need 
additional inputs from lags which are one seasonal lag (or its multiples) behind $y(t+1)$. Specifically, it would need
$x(t+1-S),y(t+1-S),x(t+1-2S),y(t+1-2S),\dots\dots\dots,
x(t+1-PS),y(t+1-PS)$. Generalizing this to a $K+1$-step ahead one-shot predictive model, we obtain 
 a model or map $F(.)$ which  predicts multi-step vector targets in  presence of exogenous variables incorporating
 stochastic seasonal
 correlations. 
% \vspace{-0.1in}
\begin{align}
	&\Big(y(t),y(t+1),y(t+2),\dots,y(t+K) \Big) = \nonumber \\
	&{F\left(\dotuline{x(t),x(t+1),x(t+2),\dots\dots\dots\dots\dots,x(t+K),} \right.} \nonumber \\
&{\dashuline{x(t-1),y(t-1),x(t-2),\dots\dots\dots, x(t-p),y(t-p),}} \nonumber \\
	& { \underbrace{x(t-S),y(t-S),\dots\dots\dots,x(t-PS),y(t-PS),} } \nonumber \\ 
	& { \underbrace{x(t+1-S),y(t+1-S),\dots,y(t+1-PS),}\dots\dots, } \nonumber \\ 
	& { \underbrace{x(t+K-S),y(t+K-S),\dots\dots,y(t+K-PS),} } \nonumber \\ 
	& {\underline{x(t-\scr{S}-1),y(t-\scr{S}-1),\dots\dots,y(t-\scr{S}-Q_1),} } \nonumber \\
	& {\underline{x(t-2\scr{S}-1),y(t-2\scr{S}-1),\dots\dots y(t-2\scr{S}-Q_2),}\dots\dots,	 }   \nonumber \\ 
	& {\left. \underline{x(t-P\scr{S}-1),y(t-P\scr{S}-1),\dots\dots,y(t-P\scr{S}-Q_{P})}\right) } 
    \label{eq:ASNARXMulti-StepGen}
\end{align}
For each additional component $y(t+i)$ in the target, there is an additional group of $2P$ inputs  $x(t+i-S),y(t+i-S),\dots\dots\dots,
x(t+i-PS),y(t+i-PS)$ that need to be added as inputs. Since $k$ varies from $0$ to $K$, we have $K$ such additional groups each of size of
$2P$. These additional groups of inputs can all be viewed as belonging to a generalized category (b), introduced
earlier. This is indicated by the additional groups of inputs introduced in
eqn.~(\ref{eq:ASNARXMulti-StepGen}) over eqn.~(\ref{eq:ASNARXGen}). All these additional $K$ groups are indicated using horizontal curly braces 
used earlier for inputs of category (b).  Note that an additional category of the future exogenous inputs 
$x(t),x(t+1),x(t+2),\dots\dots,x(t+K)$ grouped with a dotted underline additionally appears in eqn.~(\ref{eq:ASNARXMulti-StepGen}).

%\subsection{\bf ED Architecture for Multi-step Seasonal NARX Model} 
\subsection{\bf Encoder-Decoder RNN Architecture for Multi-step Seasonal NARX Model} 
\label{sec:EDMult}
\begin{figure*}[htbp]
\center
%\includegraphics{../../SIAM/SDMFiles/PlotsSDM/AAAI19FigData/EKFVsPFMAPE.pdf}
%\begin{tabular}{cc}
 %\includegraphics[width=7.0in, height=6.0in]{../../../../SampleFiguresInkScape/Retail/SeasonalityMultiStepPaper.png}
 \includegraphics[width=7.0in, height=6.0in]{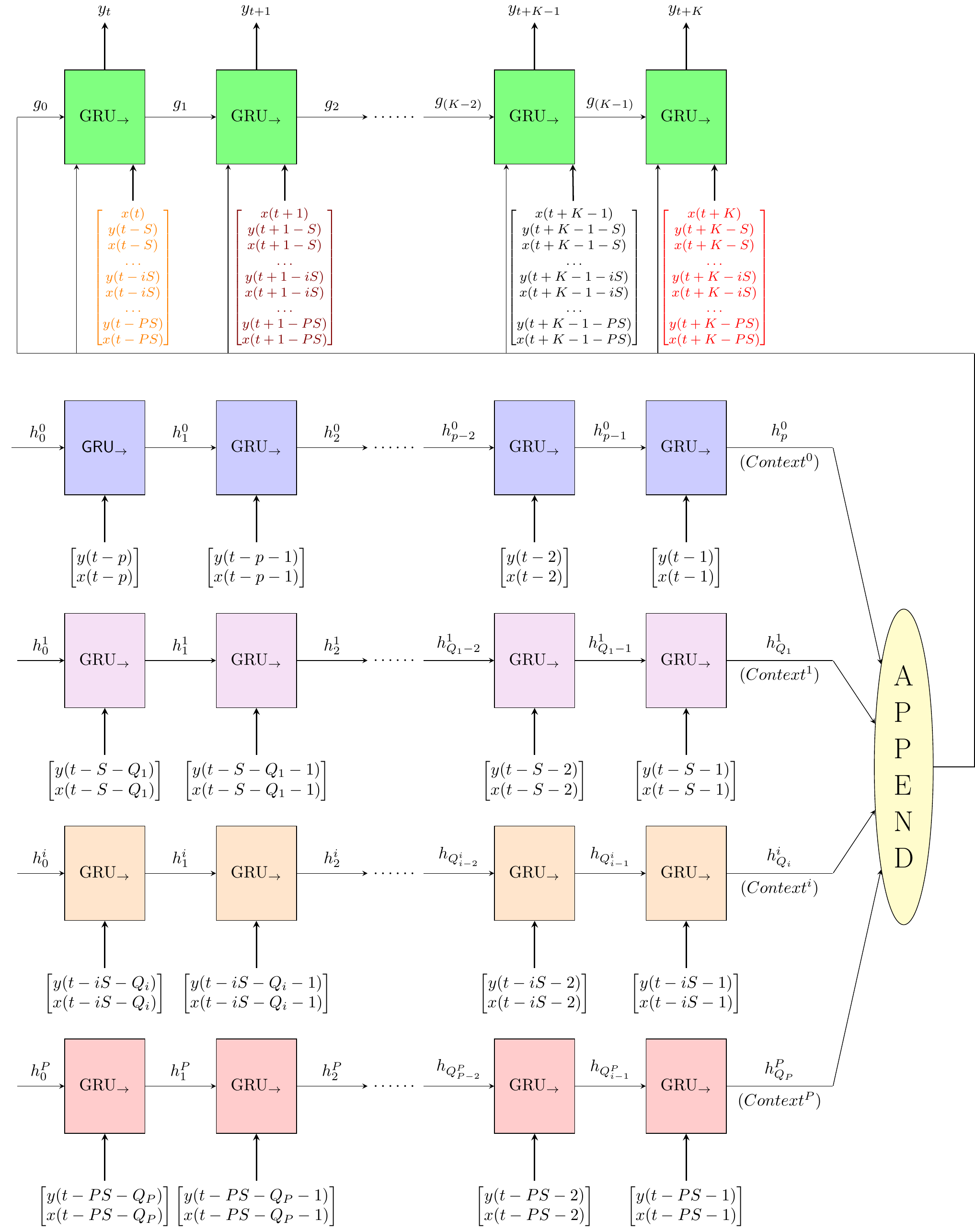}
%\subfigure[RunTime Comparison]{\includegraphics[width=1.6in,height=1.2in]{../SIAM/SDMFiles/PlotsSDM/AAAI19FigData/EKFVsPFRunTime.pdf} \label{fig:RunTimeEKFvsPF}}
%\end{tabular}
\caption{Encoder-Decoder RNN based Seasonal Multi-step NARX Architecture}
\label{fig:SeasArch}
\vspace{-0.10in}
\end{figure*}

To predict accurately during multi-step prediction, we train with vector-valued targets, the vector size equal to 
prediction
horizon. 
%The classical Seq2Seq (ED) used for machine translation can inherently tackle input-output pairs of variable sizes. 
The classical Seq2Seq (ED)
can be neatly 
adapted to the multi-step 
context where the decoder is unfolded as much as the prediction horizon length ($K+1$ steps).  

Fig.~\ref{fig:SeasMS} gives an example sequence shown in four rows, where the sequence starts from the top left and ends on the bottom
right. It gives a  pictorial view of the inputs utilized (up-to $P$ cycles behind the current
time $t$) for prediction from time $t$ on-wards. Fig.~\ref{fig:SeasArch} describes
the associated proposed architecture with color coding of inputs and blocks matched with that of Fig.~\ref{fig:SeasMS}.

We  implement eqn.~(\ref{eq:ASNARXMulti-StepGen}) in an intelligent and non-redundant fashion using a novel
ED architecture.  
We rewrite eqn.~\ref{eq:ASNARXMulti-StepGen} by reorganizing its inputs as follows which aids us in clearly associating the inputs and outputs of the above multi-step 
seasonal NARX model to the proposed ED architecture.   

\begin{align}
	&\Big(y(t),y(t+1),y(t+2),\dots,y(t+k) \Big) = \nonumber \\
	&{F\left(\dashuline{x(t-1),y(t-1),x(t-2),\dots\dots\dots, x(t-p),y(t-p),} \right.} \nonumber \\
	& {\underline{x(t-\scr{S}-1),y(t-\scr{S}-1),\dots,x(t-\scr{S}-Q_1),y(t-\scr{S}-Q_1),} } \nonumber \\
	& {\underline{x(t-2\scr{S}-1),y(t-2\scr{S}-1),\dots\dots y(t-2\scr{S}-Q_2),}\dots\dots\dots	 }   \nonumber \\ 
& {\underline{x(t-P\scr{S}-1),y(t-P\scr{S}-1),\dots,y(t-P\scr{S}-Q_{P})},}   \nonumber \\
%&{\dotuline{x(t),x(t+1),x(t+2),\dots\dots\dots\dots\dots,x(t+k),}} \nonumber \\
	& { \underbrace{x(t),x(t-S),y(t-S),\dots\dots\dots,x(t-PS),y(t-PS),} } \nonumber \\ 
	& {\underbrace{x(t+1),x(t+1-S),y(t+1-S),\dots,y(t+1-PS)},\dots,} \nonumber \\ 
	& {\left. \underbrace{x(t+K),x(t+K-S),y(t+K-S),\dots,y(t+K-PS)} \right)}  
    \label{eq:ASNARXMulti-StepGen-ED} \vspace{-0mm}
    \end{align} 
 The finer splits in inputs in Fig.~\ref{fig:SeasMS} are in sync with the input 
rearrangement of eqn.~(\ref{eq:ASNARXMulti-StepGen-ED}).    

The first group in the rearrangement of eqn.~(\ref{eq:ASNARXMulti-StepGen-ED}) comes from  standard immediate consecutive lags of order $p$. 
Note the lags shaded in blue and grouped as ``standard auto-regressive" lags in Fig.~\ref{fig:SeasMS}. These inputs of category (a) are fed
as input to Encoder $0$ (GRU units shaded in blue). This is followed by $P$ groups of
seasonal lags, where the $i^{th}$ such group is a bunch of $Q_i$ consecutive lags starting exactly $iS$ lags behind  current time $t$.
All time points in Fig.~\ref{fig:SeasMS} above the "standard auto-regressive lags`` represent these $P$ groups. Each group
here is colored differently and is fed into a separate encoder (Fig.~\ref{fig:SeasArch}) with GRU units colored in sync with the color of the associated time points  in
Fig.~\ref{fig:SeasMS}.     
%From the ED perspective, these $P$ groups of  seasonal consecutive lags are fed into $P$ separate  encoders. 
In essence, we propose to have multiple encoders depending on the order $P$ of the model. {\em This ensures that there is equal emphasis  
from seasonal lags of all orders during model building.}
As observed, we also feed all the associated past exogenous inputs/observations at these various encoder time steps.
{\em Context vectors obtained at 
the last time-step of each of these $P$ encoders are appended before feeding further as 
initial state at the decoder's first time-step.} To ensure better learning, the appended context vector can also be additionally fed as an 
input at each time step of the decoder.

{\em In addition, the ED framework can admit exogenous
inputs during $(K+1)$-step forecast horizon 
 as additional inputs at the respective time-steps of the decoder.} Let us return to the rearrangement of inputs in
 eqn.~(\ref{eq:ASNARXMulti-StepGen-ED}) from eqn.~(\ref{eq:ASNARXMulti-StepGen}).  We  observe that the future exogenous inputs
 $x(t),x(t+1),\dots,x(t+K)$ were distinctly present as the first group of inputs in  eqn.~(\ref{eq:ASNARXMulti-StepGen}). For each of these
 future exogenous inputs in eqn.~(\ref{eq:ASNARXMulti-StepGen}), there exists a unique group of $2P$ past inputs/observations (endogenous 
 + exogenous)
 $x(t+k-S),y(t+k-S),\dots,x(t+k-PS),y(t+k-PS)$, underlined with horizontal curly braces in eqn.~(\ref{eq:ASNARXMulti-StepGen}). Eqn.~(\ref{eq:ASNARXMulti-StepGen-ED}) merges each
 $x(t+k), k=0,1,\dots,K$ with its uniquely associated group as just described above. Therefore we have $K+1$ such groups of $2P+1$ inputs.
 Each of these groups is indicated in Fig.~\ref{fig:SeasMS} with a yellow shaded vertical grouping enclosed in blue dotted rectangles. The associated inputs of each group is
 color coded differently and the same color is carried forward in the decoder inputs of Fig.~\ref{fig:SeasArch}. Specifically, the $k^{th}$ such group of $2P+1$
 inputs is fed as input to the $k^{th}$ time-step of the decoder (Fig.~\ref{fig:SeasArch}). The $K+1$ outputs in 
 eqn.~(\ref{eq:ASNARXMulti-StepGen}) are the targets at the $K+1$ time-steps of the decoder output.

\comment{
{\em Overall the seasonal lag inputs  
are intelligently split 
between  encoder and decoder inputs avoiding any redundant inputs.} Lags that influence all  instants in the prediction horizon are fed via the encoders, while  lags which are exactly a cycle (or its integral multiple) behind a point in the predictive horizon are fed at the decoder in a synchronous fashion. For more details around this, refer to appendix. 
}

\subsubsection{\bf Seasonal lag distribution between encoder \& decoder}
%{\bf Seasonal lag distribution between encoder \& decoder:} 
Let us start with $P=1$ case. To predict for $y(t)$, recall that the one-step 
endogenous  model  eqn.~(\ref{eq:ASARGen}) needs as input $y(t-S)$  from exactly one cycle behind and
$Q_1$ lags preceding it.  A $(K+1)$-step ahead predictive model (eqn.~\ref{eq:ASNARXMulti-StepGen-ED} without exogenous inputs) from $y(t)$ to $y(t+K)$ would depend on data from  $y(t-S)$ to $y(t-S+k)$ (exactly one cycle behind the prediction instants)
and
$Q_1$ lags preceding $y(t-S)$. {\em Hence, note
that this block of $Q_1$ lags are invariant to the multi-step prediction horizon length. That's why this block of lags is fed as a separate encoder. Its immediately succeeding
lags are fed at the decoder end as they can be exactly matched (OR synchronized) with one of the multi-step targets  exactly one period ahead.} Thats why we
see data from  time points $y(t-S)$ to $y(t-S+k)$ fed as decoder inputs from the  first to the $(K+1)^{th}$ time-step respectively. 
For $P>1$, this kind of synchronized inputs at the
decoder can additionally come from lags exactly $2\scr{S},\dots P\scr{S}$ time points behind. This means depending on $P$, each time step of the decoder receives
input from $P$ time points each of which are exactly $i\scr{S}$ steps behind, where $i=1,\dots,P$.

\begin{remark}
{\em Overall the seasonal lag inputs  
are intelligently split 
between the encoder and decoder inputs avoiding any redundant inputs. Lags that influence all  instants in the prediction horizon are fed via the encoders, while  lags which are exactly a cycle (or its integral multiple) behind a point in the predictive horizon are fed at the decoder in a synchronous fashion.}
{\em Even in the absence of exogenous
 inputs, the ED architecture proposed still holds with multiple encoders and the decoder inputs coming from the synchronized past
 observations of the endogenous variable.}
\end{remark}

\begin{remark}
{\em We adopt a moving window approach in this paper and form a 
training example from every possible window in the time series.} An example input-output window is best illustrated in Fig.~\ref{fig:SeasMS}.
\end{remark}

\section{\bf Training and prediction on multiple sequences}
\label{sec:GreedyRec}

\begin{algorithm}[!t]
\caption{Build one or more background models to cover all sequences.}
\label{algo:BM}
\linesnumbered
%\footnotesize
	\KwIn{Set of sequences $\scr{T}$.}
	\KwOut{a set of models $\scr{M}$  which cover all sequences.}
%Initialize $\Fl=1$\; %$\CS = \Gamma$, $\CSt = \alpha_s$, $\fs = 1$, $\scr{P}=$ Set of particles grown up to $t$\;
	Perform a sequence specific scaling (min-max normalization) of each sequence (time-series) $T$ (both $Y$ and $X$ components of $T$) in $\scr{T}$.  \\
	Initialize $\scr{G}\leftarrow\scr{T}$. \\
{\em Model-Recursive-fn($\scr{G}$)}\;
%	Form training examples from all the sequences in $G$ and build one model $M_1$ using all examples, which can explain all $t \in \scr{T}$. \\
\end{algorithm}

\begin{algorithm}[!t]
\caption{{\em Model-Recursive-fn($\scr{G}$)}}
\label{algo:RecFunc}
\linesnumbered
%\footnotesize
	\KwIn{$\scr{G}$ (subset of sequences from $\scr{T}$)}%or the dimension of each $\B{z_i}$. }
	%\KwOut{a set of models $\scr{M}$  which cover all sequences.}
%Initialize $\Fl=1$\; %$\CS = \Gamma$, $\CSt = \alpha_s$, $\fs = 1$, $\scr{P}=$ Set of particles grown up to $t$\;
Form training examples from all the sequences in $G$\;
	Build one RNN model $M$ using the above possible examples, which can predict for all $T \in \scr{G}$. ($M$ predicts for all sequences in the 
	normalized domain). \;
Apply inverse sequence specific scaling for each sequence to evaluate sequence specific prediction errors\;
	Form $G_1 := \{T \in \scr{G} : e(T)\leq E_{th}\}$\;
\If{$G_1 \neq \phi$}{
	Add $(M,\scr{G}_1)$ to $\scr{M}$\;
	Form $\scr{G}_2 \lA \scr{G}\setminus\scr{G}_1$\; 
	%$\scr{G} \lA G_2$\;
	\If{$\scr{G}_2 == \phi$}{
		return\;
		}
	\Else{
	{\em Model-Recursive-fn($\scr{G}_2$)}\;
	}
		}
\Else{
	We build  sequence specific models to all sequences in  $\scr{G}$\;
	Add all these models to $\scr{M}$\;	
	return\;
		}
\end{algorithm}

While the proposed architecture can be used for a single time series case,  using it for multiple time-series prediction is not clear. To be 
able to adapt our architectures for prediction across {\em short length} multiple time-series,
we present a novel
greedy recursive algorithm. The overall idea of this algorithm is that there exist one or at most a handful of models which explain all the sequences 
in a scaled (or normalized) domain. This can be employed in situations where building RNN models per sequence is infeasible due to data scarcity at a sequence 
level. This could be very useful in situations where the variations in the exogenous variable is also little for every sequence.

Alg.~\ref{algo:BM} together with Alg.~\ref{algo:RecFunc}  explains the overall procedure. Line $1$ performs a seq-specific 
scaling of the both the $Y$ (endogenous) and $X$ (exogenous) components as an attempt to build common models in the normalized domain. The set of scaled sequences are now fed 
to the recursive function {\em Model-Recursive-fn(.)} (given in Alg.~\ref{algo:RecFunc}). Build a model using all sequences in $\scr{G}$ which can predict for
all sequences $T$ in $\scr{G}$ (line $2$). We evaluate the model performance sequence-wise for every $T\in \scr{G}$ (based on a separate validation set). The error metrics could be MASE or MAPE for
instance (refer to the next section for details). We keep aside all sequences $T\in \scr{G}$ on which the model $M$'s (validation) error ($e(T)$)  is below a user-defined threshold $(E_{th}$) into $\scr{G}_1$ (line $4$). 
If $G_1$ is empty (line $12$), then we build sequence specific models using linear time series OR shallow RNN models (line $13 - 15$) and
return.  If $\scr{G}_1$ is non-empty (line $5$), we first
add the current model $M$ and the current set of sequences $\scr{G}_1$ to $\scr{M}$ (line $6$). While
its compliment in $\scr{G}$, namely $\scr{G}_2$ (line $7$) is also non-empty (line $10$), we attempt to build an additional model greedily on $\scr{G}_2$ (set of sequences on which the
current model $M$ performs poorly), as per line $11$. If $\scr{G}_2$ is empty, we return (line $9$).

\section{\bf Results}
\label{sec:Results}
%Sym MAPE resolves the the asymmetry when True and Actual and interchanged. In that process it disturbs the symmetry that MAPE preserves for equal positive and
%negative errors. Symm MAPE also ensures the error metric is bounded between 0 and 200. For a true sales of 0, any non-zero prediction results in 200% error.

We first describe the data sets used for testing, followed by  error metrics and hyper-parameters for evaluation, and  performance results in comparison to some strong base lines.
\subsection{\bf Data Sets}
\begin{itemize}
\item {\bf D1:} This is demand data from  National Electricity Market of Australia.  Australian geography is split into $5$ disjoint regions, which means 
	we have five   power demand time series including an aggregate temperature time-series (exogenous input) in each of these regions. This is a single time-series data-set consisting of $5$ independent  time-series. D1 is $3$ months of summer data (Dec to Feb) from   these $5$ regions.  The granularity of the time-series here is half-hourly. The last $2$ weeks were used for testing.
\item{\bf D2:} M5 data-set\footnote{https://www.kaggle.com/c/m5-forecasting-accuracy/data} is a publicly available 
%\footnote{https://www.kaggle.com/c/m5-forecasting-accuracy/data}
	data-set from Walmart and contains the unit sales of different products on a daily basis spanning 5.4 years. This data is distributed across $12$ different aggregation levels. Amongst these levels,  aggregation level $8$ contains unit sales of all products, aggregated for each store and category. Price is used as exogenous input here. {\em This level contains $30$ sequences. Based on the PACF value at the $365^{th}$ (S=365) lag, we choose top 3 sequences, which we refer to as D2.} We focus on sequences which have significant PACF 
based evidence of stochastic seasonal correlations.
	\item {\bf D3:} This is  publicly available  from Walmart\footnote{https://www.kaggle.com/c/walmart-recruiting-store-sales-forecasting/data}. The measurements are weekly sales at a department level of multiple departments across 45 Walmart 
		stores. In addition to sales, there are other related measurements like CPI (consumer price index), mark-down price etc. which we use as 
		exogenous variables for weekly sales prediction. The data is collected across $3$ years and its a multiple time-series data. The whole data set consists of $2628$ 
		sequences. For purposes of this paper, {\em we ranked sequences based on total variation of  the sales and considered the top $20\%$ of
		the sequences (denoted as D3) for testing}. We essentially pick the hardest sequences which exhibit sufficient variation. The total 
variation of a $T$ length sequence $x$ is
defined as
\vspace{-0.1in}
\begin{equation}
	TV = \sum_{i=2}^{T} |(x(i+1) - x(i)|
\end{equation}
	
	\item{\bf D4:}  Data at level $10$ from M5 data-set(described above) contains unit sales  product-wise aggregated for all store/states. This level contains a total of 3049 sequences as there are that many products. In order to simulate a situation where per sequence data is relatively less,  we consider only last $3$ years  data  at a weekly resolution (instead of daily) by further aggregating sale units across a week. A single price is the exogenous variable here. 
	%In case of any price variation across the week, a weighted average price is calculated for each week (treated as the exogenous variable). 
	Further we ranked the 3049 sequences based on total variation and  considered the top 20\% of these sequences (similar to D3), which we refer to as D4 (containing 609 sequences in total). D4, product level data (level 10 - M5) has characteristics
different from D3 (Walmart), which is department level sales (aggregated across products). 
\end{itemize}

		%We test both architectures on D1 and  employ Alg.~\ref{algo:BM} towards this. We test the seasonal architecture  on the $3$ single time-series of D2, while the  'Without imputation' architecture is tested  on D3 using Alg.~\ref{algo:BM}.

\subsection{\bf Error metrics and Hyper-parameter choices}
We consider the following two  error metrics: (1){\bf MAPE} (Mean Absolute Percentage Error) (2){\bf MASE} (Mean Absolute Scale Error\cite{Hyndman06})
\comment{
\begin{itemize}
	\item {\bf MAPE} (Mean Absolute Percentage Error)
	\item {\bf MASE} (Mean Absolute Scale Error\cite{Hyndman06})
\end{itemize}
}

 The APE
is essentially relative error (RE) expressed in percentage. If $\widehat{X}$ is predicted value, while $X$ is the true value, the RE = $(\widehat{X} - X)/X$. In
the current multi-step setting, APE is computed for each step and is averaged over all steps to obtain the MAPE for one window of the prediction horizon.
The APE while has the advantage of being  a scale independent metric, can assume abnormally high values and can be misleading when the true value is very low. An
alternative complementary error metric which is scale-free could be MASE. 

The MASE is computed with reference to a baseline metric. The choice of baseline is
typically the copy previous predictor, which just replicates the previous observed value as the prediction for the next step.  
For a given window of one prediction horizon of $K$ steps ahead, 
let us denote the $i^{th}$ step error by $|\widehat{X}_i - X_i|$. The $i^{th}$ scaled error is
defined as 
\vspace{-0.1in}
\begin{equation}
	e_s^i =  \frac{|\widehat{X}_i - X_i|} { \frac{1}{n-K}\sum_{j=K+1}^{n}|X_j - X_{j-K}|}
\end{equation}
where $n$ is  no. of  points in the training set. The normalizing factor is essentially the average $i^{th}$ step-ahead error of the copy-previous baseline 
on the training set. Hence  MASE on a multi-step prediction window $w$ of size $K$ will be 
\vspace{-0.1in}
\begin{equation}
	MASE(w,K) = \frac{1}{K}\sum_{j=1}^{K} e_s^j
\end{equation}

\subsubsection{\bf Hyper-parameters}
%\label{hyper}
%{\bf Hyperparameters:} 
Tab.~\ref{tab:HP} describes the broad choice of hyper-parameters during training in our experiments. 
%We have used the same seed initialization for the proposed method and different baselines we bench-marked against.
\begin{table}[!htbp]
%\begin{center}
\caption{Model parameters for during training.}
\label{tab:HP}
\centering
\begin{tabular}{|c |c| }
 \hline
	{\bf Parameters} & {\bf Description} \\ \hline
 Batch size & 256/64   \\ %\hline
 Learning Rate & 0.002 \\ %\hline
 No. of Epochs & 40/70 \\ %\hline
 Number of Hidden layers & 1/2   \\ %\hline
 Hidden vector dimensionality  & 7/17    \\ %\hline
 Optimizer & RMSProp    \\ \hline
\end{tabular}
%\end{center}
\end{table}

\subsection{\bf Baselines}
\label{sec:SA}
We denote our seasonal approach compactly by SEDX. We focused on base-lining against traditional time-series and state-of-art RNN approaches. 
We didn't consider non-RNN DL approaches like N-Beats, Temporal Convolutional Networks etc. whose underlying
architectures are a bit different.
The baselines we benchmark our method against are as follows:
\begin{enumerate}
\item SARX - Seasonal AR with exogenous inputs (strong standard linear time-series baseline). We stick to an AR model here as a sufficiently long AR model can approximate
any ARMA model in general.   For D1, D2 the AR orders are also determined from PACF. Given
		half-an-hour granularity of the data, we choose $S = 48$ to capture daily seasonality in D1. In D2, which is daily
		data, we choose $S=365$
		to capture yearly seasonality. For D3, D4 we read off the sequence-specific orders $p$ from the respective PACFs,
		$S=52$ to capture yearly seasonality. For D3 and D4, we choose 
$P=1$ (eqn. \ref{eq:MSAR}) not only for SARX, but all the other non-linear baselines, as we have only about 3 years of data per sequence.
	\item BEDX - Basic Encoder Decoder (with only one encoder capturing the immediate lags), while the exogenous inputs of the prediction instants are fed as
		inputs to the decoder as considered in \cite{Wen17}. It is a simplified SEDX with all structures and inputs from the seasonal lags excluded.
	%\item BED - BEDX without exogenous inputs.
	%\item MTO - Many ($L$ steps) to one architecture predicting one-step ahead, where  $L$ previous lags of $x$ , $L-1$ previous lags of $Y$
	%	and the current value of $Y$  are fed as input at the $L$ steps.
	%\item MTO - Many ($L$ steps) to one architecture with no exogenous input (predicting K-steps ahead), where  all $K$ outputs are placed at the last time-step. 
	\item DeepAR \cite{Flunkert17} - explained earlier in Sec. \ref{sec:SeasEDSurvey}. We use a squared error loss for training which is equivalent to a Gaussian conditional log-likelihood of the targets
(with a constant variance). The input window length is chosen consistent with that of SEDX.
%Input window length = *   
\item LSTNet \cite{Lai18} - explained earlier in Sec. \ref{sec:SeasEDSurvey}.  Other parameters include history length = 60,No of filters = 16, AR window size = 14, skip length = 52, dropout rate = 0.3.
\item PSATT \cite{Yagmur17} - explained earlier in Sec. \ref{sec:SeasEDSurvey}. We only consider the first variant of position-based attention where each state component is weighted
	same. The second variant involving state component dependent weights  can lead to too many extra parameters. Its multivariate extension does not scale well for large
		dimensions due to increased number of parameters to be learnt. In particular, data-sets D3 and D4 have a few hundred sequences (hence a high input dimension), while
		the per-sequence length is limited. Hence we do not consider PSATT for bench-marking on D3, D4. For D1 $\&$  D2, input window length is chosen to accommodate up-to $PS$ lags.
%For History length = 60, No. of filters, etc. == * 
\end{enumerate}

In both D1 and D2, we didn't observe any evidence for integrating type non-stationarity in the ACF plots for instance. Typically, a very slow decay in the ACF is indicative of a
possible integrating type non-stationarity in the data. Similarly a slow decay at the seasonal lags is indicative of a need for seasonal differencing to cancel the random walk
non-stationarity.   

\subsubsection{\bf Assessing significance of mean error differences statistically} 
We have conducted a Welch t-test (unequal variance) based significance assessment (across all relevant
experiments) under both the mean metric (MASE, MAPE)
differences (SEDX vs Baseline) with a significance level of 0.05 for
null hypothesis rejection. In all subsequent tables, a simple way to indicate the best performing method is to  highlight in bold 
the best/least MASE/MAPE. Our test of significance can strengthen this visualization by allowing for
MASE/MAPE highlighting of other methods (multiple)  whose mean error difference with the best method's MASE/MAPE (need not be SEDX always) 
is statistically insignificant. We test 
significance of mean differences when averaged at the (finest) test
example level. 

\subsection{\bf Results on Single-Sequence Data-sets (D1,D2)}
\subsubsection{\bf Results on D1}
For D1, the multi-step horizon was chosen to be $48$ to be able to predict a day ahead (each time point is half-hourly demand). {\em There was 
evidence for seasonal correlations in the ACF in terms of local maxima at lags $48$ and $96$, which prompted us to choose $S=48$ and seasonal order 
($P=2$).} To choose the length of the associated encoders, we look to the significant PACF values just behind the $48^{th}$ and $96^{th}$
lags. Tab.~\ref{tab:NEMSeas} indicates the error metrics in comparison to $4$ feasible baselines (LSTNet is not applicable for the
single sequence case).  Our results demonstrate superior performance of SEDX against improvements of up to $1.94$ in MASE and $34\%$ in MAPE.
In particular, SEDX outperforms SARX.
Our significance analysis reveals SEDX’s MASE/MAPE reduction
compared to BEDX (in-spite of visually close MAPEs) is actually statistically
significant for R1 to R4.
%\begin{table}[!htbp]
%\vspace{-0.00in}
%%\begin{center}
%	\caption{(MASE, MAPE) across five regions in Australia}
%\label{tab:NEMSeas}
%%\footnotesize
%\centering
%%\begin{tabular}{|c ||c|c|c|c|c| }
%\begin{tabular}{|c ||c|c|c|c|c| }
% \hline
%	%& \multicolumn{3}{|c|}{SEDX better} & \multicolumn{3}{|c|}{Baseline better}\\ \cline{2-7}
	%Method  & R1 & R2 &  R3 & R4 & R5  \\  \hline
%Method  & R1 & R2 &  R3 & R4 & R5  \\  \hline
%	SEDX & {\bf (0.38,6)} & {\bf (0.37,4)}  &{\bf (0.64,4)} &{\bf(0.71,10)}  & (0.58,11) \\ %\hline
%	SARX & (0.86,15) & (0.41,5)  &(1.00,8) &(1.58,24) & (1.10,21)   \\ 
%	BEDX &(0.58,8) & (0.46,5)  &(0.69,5) &(0.73,11) &{\bf(0.55,10)}  \\ %\hline
%	DeepAR &(1.32,16) & (1.18,18)  &(1.32,9) &(1.48,18) &(2.52,45)  \\ \hline

%\end{tabular}
%%\end{center}
%\end{table}

\begin{table}[!htbp]
\vspace{-0.15in}
%\begin{center}
	\caption{(MASE, MAPE) across five regions in Australia}
\label{tab:NEMSeas}
%\footnotesize
\centering
%\begin{tabular}{|c ||c|c|c|c|c| }
\begin{tabular}{|c ||c|c|c|c|c|c| }
 \hline
	%& \multicolumn{3}{|c|}{SEDX better} & \multicolumn{3}{|c|}{Baseline better}\\ \cline{2-7}
	%Method  & R1 & R2 &  R3 & R4 & R5  \\  \hline
	Region  & SEDX & SARX &  BEDX &DeepAR  &PSATT \\  \hline
	R1 &\bf{(0.38,6)}  &(0.86,15)   &(0.58,8) &(1.32,16) &(0.98,14)  \\ %\hline
	R2 &\bf{(0.37,4)} &(0.41,5) &(0.46,5) &(1.18,18) & (1.09,11)\\
	R3 &\bf{(0.64,4)}  &(1.00,8)   &(0.69,5) &(1.32,9) & (0.87,5.85)     \\ 
	R4 &\bf{(0.71,10)} &(1.58,24)   &(0.73,11) &(1.48,18) &(1.11,15.22) \\ %\hline
	R5 &(0.58,11) &(1.10,21)   &\bf{(0.55,10)} &(2.52,45)  &(0.91,15.97) \\ \hline

\end{tabular}
%\end{center}
\vspace{-0.00in}
\end{table}

\subsubsection{\bf Results on D2}
\label{resultsd3}
	For D2, the prediction horizon was set to be $28$ days ($K=28$).  A test size of $33$ days (time-points) was set aside for each sequence in D2. This means we tested for  
	$6$ windows of width $28$ on the $33$ day test set per sequence.
	{\em Here we choose the seasonal order $P = 1$ with $S=365$ (yearly seasonality is exploited) by analyzing ACF values. } 
	Tab.~\ref{tab:NEMSeas1} gives the detailed comparison with all the $4$ (single sequence) strong baselines in terms of both the error metrics. Our results demonstrate that
	SEDX mostly outperforms all $4$ baselines on all $3$ sequences, except for BEDX doing equally well (in a statistical sense) on seq 3. In particular, 
we observe improvements of up to $1.09$ in MASE and $17.4\%$ in MAPE in favor of our method.
\begin{table}[!htbp]
%\begin{center}
	\caption{(MASE,MAPE) across $3$ sequences in D2}
\label{tab:NEMSeas1}
\centering
%\begin{tabular}{|c ||c|c|c|c|c| }
\begin{tabular}{|c ||c|c|c|c|c| }
 \hline
	%& \multicolumn{3}{|c|}{SEDX better} & \multicolumn{3}{|c|}{Baseline better}\\ \cline{2-7}
	%Method  & R1 & R2 &  R3 & R4 & R5  \\  \hline
	Seq  & SEDX  &  BEDX & DeepAR & SARX & PSATT  \\  \hline
	%1 &  (0.86,10 ) &  (1.03,10 )  & (0.89,10 ) & (2.30,39 )   \\ %\hline
	\comment{
	1 & \bf{(0.38,5 )} & (0.56,7 )  &(0.49,8 ) &(1.90,45 )   \\ 
	2 &(0.90,10 ) & \bf{(0.87,9 )}  &(1.03,13 ) &(1.92,34 )   \\ %\hline
	3 &(0.90,14 ) & (0.98,14 )  &\bf{(0.85,13 )} &(1.40,24)   \\ %\hline
	4 &\bf{(0.70,12 )} & (1.23,19 )  &(0.73,13 ) &(1.03,20 )   \\ %\hline
	%6 &(0.74,14 ) & (0.83,15 )  &(0.74,16 ) &(0.72,14 )   \\ %\hline
	5 &\bf{(0.52,9 )} & (0.64,10 )  &(0.53,9 ) &(1.18,23 )   \\ %\hline
	6 &\bf{(0.64,8 )} & (1.80,23 )  &(0.74,10 ) &(0.86,14 )   \\ %\hline
	7 &\bf{(0.41,8 )} & (0.81,15 )  &(0.43,8 ) &(0.76,15 )   \\ %\hline
	8 &(0.48,7 ) & (0.55,7 )  &\bf{(0.43,6 )} &(2.41,42 )   \\ %\hline
	9 &\bf{(0.43,7 )} & (1.18,15 )  &(0.49,8 ) &(1.61,40  )  \\ %\hline
	10 &\bf{(0.52,8 )} & (1.17,17 )  &(0.53,9 ) &(1.11,16  )  \\ \hline
	4 &\bf{(0.60,8.7 )} & (0.77,11.22 )  &(0.99,15.6 ) &\bf{(0.64,7.8)}   &(0.78,11.93)\\ %\hline
	%6 &(0.74,14 ) & (0.83,15 )  &(0.74,16 ) &(0.72,14 )   \\ %\hline
	5 &\bf{(0.39,8.6 )} & \bf{(0.4,8.3 )}  &(0.93,19.7) &(0.70,13.76)  &(1.24,24.4) \\ \hline
	}
	1 & \bf{(0.62,10)} & (0.77,11.8)  &(1.39,22.2) &(1.15,17.8) &(1.71,27.4)   \\ 
	2 &\bf{(0.69,9.9)} & (0.82,11.8)  &(1.11,16.2) &(0.88,12.2) &(0.99,13.9)   \\ %\hline
	3 &\bf{(0.69,11.1)} & \bf{(0.76,11.7)}  &(1.52,23.2) &(0.91,14)   &(0.95,14.1)\\ \hline %\hline
\end{tabular}
%\end{center}
\end{table}

\subsection{\bf Results on Multiple-Sequence Data-sets (D3,D4)}
\subsubsection{\bf Results on D3}
We first demonstrate the effectiveness of SEDX on D3. A test size of $15$ weeks (time-points) was set aside for each sequence in D3. We choose $K=10$ time-steps in the decoder for
training which means we tested for $6$ windows of width $10$ on the $15$ week test set per sequence. For both D3 and D4, we append all sequences (normalized using  
sequence-specific normalization) into one long sequence and look at its Partial auto-correlation function (PACF). This enables us to fix the number of time-steps in the encoders
corresponding to the seasonal correlations, which is typically much lesser than the number of time-steps in encoder $1$ (which captures
	standard lags).  On D3, with MASE based threshold ($E_{th}$) of 0.3, {\em Model-Recursive-fn()} ran for $2$ rounds.

\comment{
\begin{table}[!htbp]
%\begin{center}
\caption{Percentage of sequences where SEDX does better.}
\label{tab:FracSl}
\centering
\begin{tabular}{|c ||c|c| }
 \hline
 Baseline &MASE based & MAPE based   \\ \hline
 BEDX &79 & 79   \\ %\hline
 BED & 78 & 79   \\ %\hline
 MTO & 80 & 80    \\ %\hline
 SARMAX & 62 & 61    \\ \hline
\end{tabular}
%\end{center}
\end{table}
}
\begin{figure}[!htbp]
\center
	 \includegraphics[width=3.2in,height=1.5in]{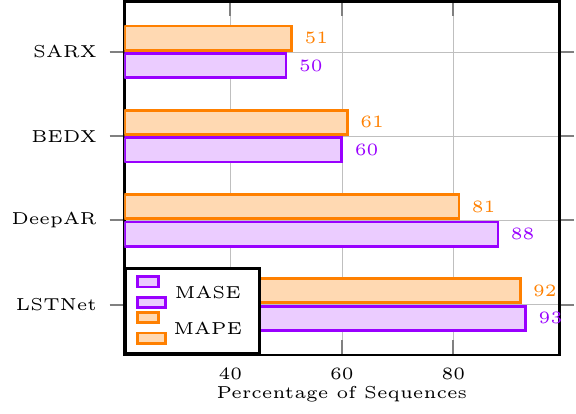}
%\begin{tabular}{cc}
%\subfigure[12 week data]{\includegraphics[width=2.50in,height=1.5in]{../../SIAM/SDMFiles/PlotsSDM/AAAI19FigData/MultiStepMAENew3.pdf} \label{fig:MultStepNew}} 
%\subfigure[5 week data]{\includegraphics[width=2.50in,height=1.5in]{../../SIAM/SDMFiles/PlotsSDM/AAAI19FigData/MultiStepMAEOld3.pdf} \label{fig:MultStepOld}}
%\end{tabular}
\caption{Percentage of Sequences where SEDX does better}
	 \label{fig:FracSl}
\vspace{-0.1in}
\end{figure}
\begin{table}[!htbp]
\vspace{-0.00in}
\caption{Max, Avg \& Min of MASE \& MAPE across all sequences}
\label{tab:MaxAvgMinSeas}
\footnotesize
\centering
\begin{tabular}{|c ||c|c|c|c|c|c| }
 \hline
	& \multicolumn{3}{|c|}{MASE based} & \multicolumn{3}{|c|}{MAPE based in \%}\\ \cline{2-7}
	Method  & Max	& Avg  &  Min & Max	& Avg  &  Min  \\  \hline
	SEDX &2.75 & {\bf0.54}   &0.16 & 76 &{\bf13} & 2  \\ %\hline
 SARX &2.38  &0.56   &0.11 &69 &14 &2   \\ 
 BEDX &2.45 &0.65   &0.13 &265 &19 &2  \\ %\hline
 DeepAR &7.58  &1.32   &0.27 &278 &28 &3  \\ %\hline
 %MTOX &3.61  &0.94    &0.19 &286 &32 &3  \\ %\hline
 LSTNet &5.48  & 1.39   &0.24 &156 &32 & 3 \\ %\hline
\hline
\end{tabular}
%\end{center}
\vspace{-0.000in}
\end{table}
%SEDX &2.65 & {\bf 0.52}  &0.1 & 72 &{\bf 12} & 2 \\ %\hline
Fig.~\ref{fig:FracSl}  gives a detailed breakup of percentage of sequences on which SEDX did better compared to the $4$ baselines. It demonstrates {\em SEDX does better on at least
$50\%$ and up to $80\%$ of the sequences compared to all considered baselines.}

Tab.~\ref{tab:MaxAvgMinSeas} gives average, max and min across sequences (of MASE and MAPE) for all methods.  It demonstrates that on an average SEDX does better than all
baselines based on both  complementary metrics. {\em MASE improvements are up to $0.85$ while the MAPE improvements are up to $19\%$.}
Min and max were
provided in Tab.~\ref{tab:MaxAvgMinSeas} (and Tab.~\ref{tab:MaxAvgMinSeasm5mase3new})  to honestly gauge error spread limits
across sequences in D3 and D4. Viewing them as metric can be
misleading.

On D3,  SEDX performance might appear similar to
SARX. But note from Fig.~\ref{fig:FracSl} and Tab.~\ref{tab:MaxAvgMinSeas}, that SEDX
is actually complementing SARX on the 526 sequences of D1.
SEDX is doing better than SARX on $50\%$ of the sequences and
giving a statistically significant improvement  of $0.2$(MASE)
and $6\%$ (MAPE) on sequences where SEDX does better (Tab.~\ref{tab:CondAvgMASESeas},\ref{tab:CondAvgMAPESeas}).

Tab.~\ref{tab:CondAvgMASESeas} looks at the (conditional) average MASE under two conditions with respect to each baseline: (i)average over those sequences on 
which SEDX fares better, (ii)average over those sequences on which the baseline does better.  At this level, MASE improvements of at least $0.20$ while up to $0.93$ are observed.  
Tab.~\ref{tab:CondAvgMAPESeas} considers a similar  (conditional) average MAPE. At this level of MAPE, there are improvements of at least $6\%$ to up to $20\%$.
\begin{table}[!htbp]
\vspace{-0.0in}
\caption{Avg MASE when (i)SEDX is better (ii)Baseline is better.}
\label{tab:CondAvgMASESeas}
\footnotesize
\centering
\begin{tabular}{|c ||c|c|c|c|c|c| }
 \hline
	& \multicolumn{3}{|c|}{SEDX better} & \multicolumn{3}{|c|}{Baseline better}\\ \cline{2-7}
	Method  & SEDX & BLine &  Diff & SEDX & BLine &  Diff \\  \hline
 %SEDX &25 & 15  &25 & 15&25 & 15 \\ \hline
	SARX &0.47  &0.67   &{\bf0.20} &0.61  &0.46 &0.15   \\ 
	BEDX &0.49 &0.79   &{\bf0.30} &0.62  &0.44 &0.18  \\ %\hline
	DeepAR &0.50  &1.40   &{\bf0.90} &0.86 &0.65 &0.21  \\ %\hline
	%MTOX &0.51  &0.97    &{\bf0.46} &0.81  &0.68 &0.13  \\ %\hline
	LSTNet &0.51  &1.44    &{\bf0.93} &0.98 &0.70 &0.28  \\ %\hline
\hline
\end{tabular}
%\end{center}
\vspace{-0.00in}
\end{table}
\begin{table}[!htbp]
%\vspace{-0.15in}
\caption{Avg MAPE when (i)SEDX is better (ii)Baseline is better.}
\label{tab:CondAvgMAPESeas}
\footnotesize
\centering
\begin{tabular}{|c ||c|c|c|c|c|c| }
 \hline
	& \multicolumn{3}{|c|}{SEDX better} & \multicolumn{3}{|c|}{Baseline better}\\ \cline{2-7}
	Method  & SEDX & BLine &  Diff & SEDX & BLine &  Diff \\  \hline
 %SEDX &25 & 15  &25 & 15&25 & 15 \\ \hline
	SARX &10  &16   &{\bf6} &18 &13 &5  \\ 
	BEDX &11 &24   &{\bf13} &18 &12 &6 \\ %\hline
	DeepAR &13  &32   &{\bf19} &17 &12 &5  \\ %\hline
	%MTOX &14  &34    &{\bf20} &14 &12 &2  \\ %\hline
	LSTNet &14  &34    &{\bf20} &18 &14 &4  \\ %\hline
\hline
\end{tabular}
\vspace{-0.00in}
%\end{center}
\end{table}

\subsubsection{\bf Results on D4}
\label{resultsd4}
For D4, where data is at a weekly granularity a prediction horizon of $8$ weeks ($K$ = 8) was chosen for training while the chosen test size was $11$ weeks for each sequence. This means for each sequence there are $4$ (input-output) windows of output window width $8$ on which we test.
We chose a yearly seasonality here (with $S=52$) similar to D3. On D4, with MASE based threshold ($E_{th}$) of 0.3, {\em
Model-Recursive-fn()} ran for $1$ round.  

In  this experiment, we observe a considerable fraction of sequences, where each of the $5$ methods (proposed + $4$ baselines) exhibit either high MASE or  high MAPE. To quantify  how high is unacceptable, we consider an MASE threshold of $1$ (beyond which a naive copy previous predictor would be better) and MAPE threshold of $30\%$. For these thresholds, we find $143$ sequences on which each of the methods either have an MASE $>1$ or MAPE $>30\%$. We have excluded  such sequences because neither our method nor any of the proposed baselines perform within acceptable limits on these. For such sequences, one can potentially explore simpler baselines like copy previous predictor or ARX model (without seasonality). We demonstrate results on the remaining $466$ sequences  where at least one of the $5$ methods (models) have both MASE $<1$ and MAPE $<30\%$.

%condition: (sedx mase<1 and sedx mape < 0.30) or(bedx mase < 1 and bedx mape <0.30) or(for deepar) or(for lstnet) total sequences=466

% (sedx mase >1 or sedx mape > 0.30) and (bedx mase > 1 or sedx mape > 0.30)and (for deepar) and (for lstnet) total sequences = 143

\comment{
\begin{table}[!htbp]
%\begin{center}
\caption{Percentage of sequences where SEDX does better}
\label{tab:m5mape3new}
\centering
\begin{tabular}{|c |c| c|}
 \hline
	{\bf Models} & {\bf MASE} &{\bf MAPE} \\ \hline
 SARX &59 &56  \\ %\hline
 BEDX &62&63  \\ %\hline
 DeepAR  &69 &72   \\ %\hline
 LSTNet &97 &97\\ \hline
\end{tabular}
%\end{center}
\end{table}
}
\begin{figure}[!htbp]
\center
	 \includegraphics[width=3.2in,height=1.5in]{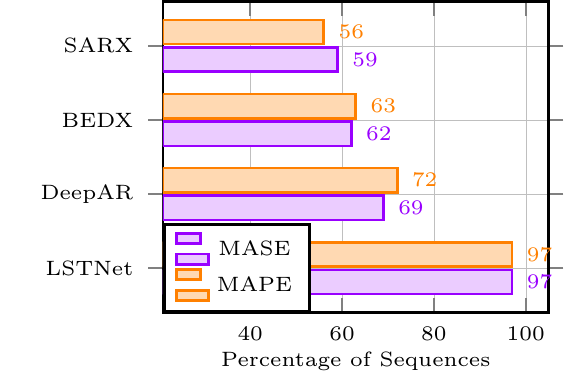}

%\begin{tabular}{cc}
%\subfigure[12 week data]{\includegraphics[width=2.50in,height=1.5in]{../../SIAM/SDMFiles/PlotsSDM/AAAI19FigData/MultiStepMAENew3.pdf} \label{fig:MultStepNew}} 
%\subfigure[5 week data]{\includegraphics[width=2.50in,height=1.5in]{../../SIAM/SDMFiles/PlotsSDM/AAAI19FigData/MultiStepMAEOld3.pdf} \label{fig:MultStepOld}}
%\end{tabular}
\caption{Percentage of Sequences when SEDX does better(pictorially)}
	 \label{fig:FracSld4}
\vspace{-0.1in}
\end{figure}

%Tab.~\ref{tab:m5mape3new} 
Fig.~\ref{fig:FracSld4} represents the percentage of (the 466) sequences on which SEDX is better compared to all four baselines in terms of both the metrics. It shows that SEDX does better on at-least 59\%(56\%) of the sequences and up-to 97\%(97\%) in terms of MASE(MAPE), compared to all baselines.

\begin{table}[!htbp]
\vspace{-0.00in}
\caption{Max, Avg and Min of MASE and MAPE across all sequences}
\label{tab:MaxAvgMinSeasm5mase3new}
\footnotesize
\centering
\begin{tabular}{|c ||c|c|c|c|c|c| }
 \hline
	& \multicolumn{3}{|c|}{MASE based} & \multicolumn{3}{|c|}{MAPE based in \%}\\ \cline{2-7}
	Method  & Max	& Avg  &  Min & Max	& Avg  &  Min  \\  \hline
	SEDX &6.45 &0.75 &0.01 &81.35 &\bf{15.42}&3.57  \\ %\hline
 SARX &3.12 &\bf{0.68} &0.16 &76.15 &16.48 &4.44  \\ 
 BEDX &6.78 &0.83 &0.01 &99.36 &17.29 &3.51  \\ %\hline
 DeepAR &6.62 &0.91  &0.02 &88.22 &19.74 &2.97  \\ %\hline
 %MTOX &3.61  &0.94    &0.19 &286 &32 &3  \\ %\hline
 LSTNet &17.29 &1.78 &0.02 &307.54 &35.08 &6.20 \\ %\hline
\hline
\end{tabular}
%\end{center}
\end{table}

Tab.~\ref{tab:MaxAvgMinSeasm5mase3new} gives the average, max and min of MASE and MAPE across  the (466) sequences for all the models. It shows that on an average SEDX is better than all considered baselines in terms of MAPE while  SARX does better in terms of MASE. %(From Tab.\ref{tab:m5mape3new}, SARX does better on 59\% sequences compared to SARX, which means for some sequences MASE value of SARX is really good.)

\begin{table}[!htbp]
\vspace{-0.0in}
\caption{Average MASE when (i)SEDX is better (ii)Baseline is better.}
\label{tab:CondAvgMASESeasnew}
\footnotesize
\centering
\begin{tabular}{|c ||c|c|c|c|c|c| }
 \hline
	& \multicolumn{3}{|c|}{SEDX better} & \multicolumn{3}{|c|}{Baseline better}\\ \cline{2-7}
	Method  & SEDX & BLine &  Diff & SEDX & BLine &  Diff \\  \hline
 %SEDX &25 & 15  &25 & 15&25 & 15 \\ \hline
	SARX &0.37  &0.78   &0.41 &1.31  &0.54 &\bf{0.77}   \\ 
	BEDX &0.73 &0.92   &\bf{0.19} &0.79  &0.68 &0.11  \\ %\hline
	DeepAR &0.69 &1.00   &\bf{0.31} &0.89 &0.71 &0.18  \\ %\hline
	%MTOX &0.51  &0.97    &{\bf0.46} &0.81  &0.68 &0.13  \\ %\hline
	LSTNet & 0.75 &1.81    &\bf{1.06} &1.01 &0.89 &0.12  \\ %\hline
\hline
\end{tabular}
%\end{center}
\vspace{-0.00in}
\end{table}
Tab.~\ref{tab:CondAvgMASESeasnew} represents the average MASE under two conditions: (i) sequences where SEDX does better (ii) sequences where baseline does better. It shows that MASE improvement of our proposed method is at-least $0.19$ and up-to $1.06$.
\begin{table}[!htbp]
\vspace{-0.00in}
\caption{Average MAPE when (i)SEDX is better (ii)Baseline is better.}
\label{tab:CondAvgMAPESeasnew}
\footnotesize
\centering
\begin{tabular}{|c ||c|c|c|c|c|c| }
 \hline
	& \multicolumn{3}{|c|}{SEDX better} & \multicolumn{3}{|c|}{Baseline better}\\ \cline{2-7}
	Method  & SEDX & BLine &  Diff & SEDX & BLine &  Diff \\  \hline
 %SEDX &25 & 15  &25 & 15&25 & 15 \\ \hline
	SARX &12.83  &18.40   &\bf{5.57} &18.69 &14.06 &4.63 \\ 
	BEDX &15.04 &19.22   &\bf{4.18} &16.06 &14.05 &2.01 \\ %\hline
	DeepAR &15.15  &22.43   &\bf{7.28} &16.11 &12.77 &3.34  \\ %\hline
	%MTOX &14  &34    &{\bf20} &14 &12 &2  \\ %\hline
	LSTNet &15.09  &35.50    &\bf{20.41} &25.98 &21.51 &4.47  \\ %\hline
\hline
\end{tabular}
\vspace{-0.0in}
%\end{center}
\end{table}

Tab.~\ref{tab:CondAvgMAPESeasnew}  also represents similar conditions based on MAPE. There are MAPE improvement of at-least $4.18\%$ and up-to $20.41\%$. Note that based on MAPE,  the average conditional improvement (indicated as Diff) of SEDX   when it does better (left half of the table) is uniformly better than the average conditional improvements of  all other baselines (Diff column on the right half of the table). 

On D4, SEDX performance
(over SARX) is actually better than in D3. Even though
overall MASE avg of 0.68 (not MAPE please note) across all
sequences looks better for SARX (Tab.~\ref{tab:MaxAvgMinSeasm5mase3new}), its only better on $41\%$
of sequences in D4 (Fig.~\ref{fig:FracSld4}). Tab.~\ref{tab:CondAvgMAPESeasnew} indicates SEDX achieves a statistically significant
improvement of $5.57\%$ (over SARX) on $56\%$ of sequences.

\section{\bf Conclusions}
\label{sec:Conc}
We proposed a novel  ED architecture for forecasting with multi-step (target) learning feature and prediction ability.
The architecture generalized a  linear multiplicative Seasonal ARX model using multiple encoders, each of which  capture correlations from one or more cycles
behind the prediction instant.  The seasonal inputs were intelligently split between encoder and decoder without redundancy. We also proposed a greedy recursive grouping algorithm 
to build background predictive models (one or at most a few)  for the multiple time
series problem 
%when the sequence lengths are small OR when the exogenous input shows minimal fluctuation in its values. 
We tested the proposed architecture and grouping algorithm  on multiple real data sets, where our proposed architecture did mostly better than all  strong baselines
while it outperformed many of them.  
%\comment{
As  future work, we would like to investigate how the proposed architecture could be utilized to better capture cross-sequence effects for multi time-series prediction. We  reckon
that the proposed architecture can be potentially useful  in many more real world scenarios. 
%}

\bibliographystyle{IEEEtran}
%\bibliography{IEEEabrv,Transportation}
\bibliography{Transportation}
\appendices

\section{Matrix Factorization approaches}
 Matrix factorization methods have been another class of methods to analyze multiple time series. Traditionally used for recommendation systems \cite{Koren09}, they have also been used for 
 analyzing 
multiple time series \cite{Xiong10,Rallapalli10}. TRMF \cite{TRMF16} is a temporally regularized matrix factorization approach capable of  prediction (unlike previous MF approaches) 
while  learning the 
temporal dependencies via an auto-regressive temporal regularizer  in a data-driven fashion. A matrix factorization approach is essentially a global
approach where dependencies across sequences are captured. An extension of TRMF is considered in DEEPGLO \cite{Sen19} which in addition to global features also considers capturing local per 
time-series characteristics. It uses a TCN for regularization (during MF) and prediction and can potentially capture (global) non-linear dependencies unlike 
TRMF. It then combines these global predictions locally using another TCN.

\end{document}